# Solving Factored MDPs with Continuous and Discrete Variables

**Carlos Guestrin**
Berkeley Research Center
Intel Corporation

**Milos Hauskrecht**
Department of Computer Science
University of Pittsburgh

**Branislav Kveton**
Intelligent Systems Program
University of Pittsburgh

## Abstract

Although many real-world stochastic planning problems are more naturally formulated by hybrid models with both discrete and continuous variables, current state-of-the-art methods cannot adequately address these problems. We present the first framework that can exploit problem structure for modeling and solving hybrid problems efficiently. We formulate these problems as hybrid Markov decision processes (MDPs with continuous and discrete state and action variables), which we assume can be represented in a factored way using a hybrid dynamic Bayesian network (hybrid DBN). This formulation also allows us to apply our methods to collaborative multiagent settings. We present a new linear program approximation method that exploits the structure of the hybrid MDP and lets us compute approximate value functions more efficiently. In particular, we describe a new factored discretization of continuous variables that avoids the exponential blow-up of traditional approaches. We provide theoretical bounds on the quality of such an approximation and on its scale-up potential. We support our theoretical arguments with experiments on a set of control problems with up to 28-dimensional continuous state space and 22-dimensional action space.

## 1 Introduction

Markov decision processes (MDPs) [2, 3] offer an elegant mathematical framework for representing sequential decision problems in the presence of uncertainty. While standard solution techniques, such as value or policy iteration, scale-up well in terms of the total number of states and actions, these techniques are less successful in real-world MDPs represented by state and action variables. In purely discrete settings, the running time of these algorithms grows exponentially in the number variables, the so called *curse of dimensionality*. Furthermore, many real-world problems include a combination of continuous and discrete state and action variables. In this hybrid setting, the continuous components are typically discretized, leading again to an exponential blow up in the number of variables.

In this work, we present the first MDP framework that can exploit problem structure to represent and solve large hybrid stochastic planning problems efficiently. We model these problems using a *hybrid factored MDP* where the dynamics are represented compactly by a probabilistic graphical model, a hybrid dynamic Bayesian network (DBN) [7]. Unfortunately, compact DBN parameterizations of hybrid MDPs do not guarantee the existence of efficient exact solutions to these problems. In hybrid settings, this issue is accentuated, as exact parametric solutions may not exist. We thus approximate the solution with a simpler parametric representation. We focus on linear approximations, where the value function is approximated by a linear combination of basis functions [1, 3]. Specifically, we use a *factored (linear) value function* [12], where each basis function depends on a small number of state variables. This architecture is central for obtaining efficient approximation algorithms for factored MDPs [12, 8].

In hybrid settings, each basis function may depend on both continuous and discrete state components. We show that the weights of this approximation can be optimized using a convex formulation we call *hybrid approximate linear programming* (HALP). The HALP reduces to the approximate linear programming (ALP) formulation [15] in purely discrete settings. When applied to discrete factored MDPs, the ALP approach leads to linear programming (LP) problems with a small number of variables, but with exponentially many constraints; one constraint per each state-action pair. Guestrin *et al.* [8] present an efficient LP decomposition technique that exploits the structure of the factored model to represent this LP by an equivalent, exponentially-smaller problem. Schuurmans and Patrascu [14] build on this decomposition technique to design an efficient constraint generation approach for the ALP problem. Alternatively, de Farias and Van Roy [5] propose a sampling approach for selecting a subset of the exponentially-large constraint set.

In purely continuous settings, the HALP reduces to the formulation recently proposed by Hauskrecht and Kveton [11] for factored continuous-state MDPs (CMDPs). This formulation requires the computation of several complex integration problems, which Hauskrecht and Kveton address by identifying conjugate classes of parametric transition models and basis functions that lead to efficient closed-form solutions. Their formulation also includes a constraint for each state, now leading to an infinite number of constraints. To solve the problem they apply the constraint sampling method in combination with various heuristics to generate a small set of constraints that define an LP approximation with no theoretical guarantees.

In this paper, we first provide theoretical guarantees for the solutions obtained by the HALP formulation that also apply to the purely continuous setting of Hauskrecht and Kveton. Specifically, we extend the theoretical analysis of de Farias and Van Roy [6] for discrete ALPs to the continuous and hybrid settings addressed by our HALP formulation, providing bounds with respect to the best approximation in the space of the basis functions.

Although theoretically sound, the HALP formulation may still be hard to solve directly. In particular, the HALP leads to a linear program with an infinite number of constraints. We address this problem by defining a relaxed formulation, $\varepsilon$-HALP, over a finite subset of constraints, using a new *factored discretization* technique that exploits structure in the continuous components. Once discretized, this formulation can be solved efficiently by existing factored ALP methods [8, 14]. We finally provide a bound on the loss in the quality of the solution of this relaxed relaxed $\varepsilon$-HALP with respect to that of the complete HALP formulation.

We illustrate the feasibility of our formulation and its solution algorithm on a sequence of control optimization problems with 28-dimensional continuous state space and 22-dimensional action space. These nontrivial dynamic optimization problems are far out of reach of classic solution techniques.

## 2 Multiagent hybrid factored MDPs

Factored MDPs [4] allow one to exploit problem structure to represent exponentially-large MDPs compactly. Here, we extend this representation to hybrid systems with continuous and discrete variables. A multiagent hybrid factored MDP is defined by a 4-tuple $(\mathbf{X}, \mathbf{A}, P, R)$ that consists of a state space $\mathbf{X}$ represented by a set of state variables $\mathbf{X} = \{X_1, \ldots X_n\}$, an action space $\mathbf{A}$ defined by a set of action variables $\mathbf{A} = \{A_1, \ldots A_m\}$, a stochastic transition model $P$ modeling the dynamics of a state conditioned on the previous state and action choice, and a reward model $R$ that quantifies the immediate payoffs associated of a state-action configuration.

**State variables:** In a hybrid model, each state variable can be either discrete or continuous. We assume that every continuous variables is bounded to a $[0, 1]$ subspace, and each discrete variable $X_i$ takes on values in some finite domain $\mathrm{Dom}(X_i)$, with $|\mathrm{Dom}(X_i)|$ possible values. A state is defined by a vector $\mathbf{x}$ of value assignments to each state variable. We split this state vector $\mathbf{x}$ into discrete and continuous components denoted by $\mathbf{x} = (\mathbf{x}_D, \mathbf{x}_C)$, respectively.

**Actions:** We assume a distributed action space, such that every action corresponds to one agent. As with state variables, the global action $\mathbf{a}$ is defined by a vector of individual action choices that can be divided into discrete $\mathbf{a}_D$ and continuous $\mathbf{a}_C$ components.

**Factored transition:** We define a state transition model using a *dynamic Bayesian network (DBN)* [7]. Let $X_i$ denote a variable at the current time and let $X'_i$ denote the same variable at the successive step. The *transition graph* of a DBN is a two-layer directed acyclic graph whose nodes are $\{X_1, \ldots, X_n, A_1, \ldots, A_m, X'_1, \ldots, X'_n\}$. We denote the parents of $X'_i$ in the graph by $\mathsf{Par}(X'_i)$. For simplicity of exposition, we assume that $\mathsf{Par}(X'_i) \subseteq \{\mathbf{X}, \mathbf{A}\}$, *i.e.*, all arcs in the DBN are between variables in consecutive time slices. Each node $X'_i$ is associated with a *conditional probability function (CPF)* $p(X'_i \mid \mathsf{Par}(X'_i))$. The transition probability $p(\mathbf{x}' \mid \mathbf{x}, \mathbf{a})$ is then defined to be $\prod_i p(x'_i \mid \mathbf{u}_i)$, where $\mathbf{u}_i$ is the value in $\{\mathbf{x}, \mathbf{a}\}$ of the variables in $\mathsf{Par}(X'_i)$. In our hybrid setting, the CPF associated with $X'_i$ is defined either in terms of a distribution or a density function depending on whether $X'_i$ is discrete or continuous.

**Parameterization of CPFs:** The transition model for each variable is local, as each CPF depends only on a small subset of state variables and on the actions of a small number of agents. In hybrid settings, we must also ensure that the local transition model can be parameterized compactly. To model the transitions of a continuous state variable in $[0, 1]$ with hybrid parents we assume beta or mixture of beta densities as proposed recently by Hauskrecht and Kveton [11]. Here, the CPF is given by:

$$p(X'_i \mid \mathsf{Par}(X'_i)) = \mathrm{Beta}(X'_i \mid h^1_i(\mathsf{Par}(X'_i)), h^2_i(\mathsf{Par}(X'_i))),$$

where $h^1_i(\mathsf{Par}(X'_i)) > 0, h^2_i(\mathsf{Par}(X'_i)) > 0$ are any positive functions of the value of the parents of $X'_i$ that define the sufficient statistics of the beta distribution. Note that the functions $h^1_i$ and $h^2_i$ can depend on both continuous and discrete parent variables $\mathsf{Par}(X'_i)$. Additionally, we allow the CPF $p(X'_i \mid \mathsf{Par}(X'_i))$ to be modeled by a (weighted) mixture of beta distributions. This mixture model provides a very general class of distributions for random variables in $[0, 1]$. Additionally, the choice of (mixture of) beta distributions allow us to perform many operations required by our algorithm very efficiently, as discussed in Section 4.

The CPF for a discrete state variable with hybrid parents can be defined using a rather general discriminant function approach. Specifically, the conditional distribution $P(X'_i \mid \mathsf{Par}(X'_i))$ can be implemented in terms of $|\mathrm{Dom}(X_i)|$ discriminant functions $d_j(\mathsf{Par}(X'_i)) > 0$. These functions specify the distribution over $X'_i$ by:

$$P(X'_i = j \mid \mathsf{Par}(X'_i)) = \frac{d_j(\mathsf{Par}(X'_i))}{\sum_{u=1}^{|\mathrm{Dom}(X_i)|} d_u(\mathsf{Par}(X'_i))}. \quad (1)$$

This discriminant function representation allows us to represent a very wide class of distributions.

**Rewards:** We assume that reward function for every state and action also decomposes as the sum of partial reward functions defined on subsets of state and action variables:

$$R(\mathbf{x}, \mathbf{a}) = \sum_j R_j(\mathbf{x}_j, \mathbf{a}_j),$$

where $R_j$ denote partial reward functions, and $\mathbf{x}_j$ and $\mathbf{a}_j$ define the restriction of the full state and action to the subspace associated with the scope of $R_j$.

**Policy:** Given our factored MDP, our objective is to find a control policy $\pi^* : \mathbf{X} \rightarrow \mathbf{A}$ that maximizes the infinite-horizon, discounted reward criterion: $E[\sum_{i=0}^{\infty} \gamma^i r_i]$, where $\gamma \in [0, 1)$ is a discount factor, and $r_i$ is a reward obtained in step $i$.

**Value function:** The value of the optimal policy satisfies the Bellman fixed point equation [2, 3]:

$$V^*(\mathbf{x}) = \sup_{\mathbf{a}} \left[ R(\mathbf{x}, \mathbf{a}) + \gamma \sum_{\mathbf{x}'_D} \int_{\mathbf{x}'_C} p(\mathbf{x}' \mid \mathbf{x}, \mathbf{a}) V^*(\mathbf{x}') \right], \quad (2)$$

where $V^*$ is the value of the optimal policy. Given the value function $V^*$, the optimal policy $\pi^*(\mathbf{x})$ is defined by the composite action $\mathbf{a}$ optimizing Eqn 2.

## 3 Approximate linear programming solutions for hybrid MDPs

There exist a number algorithms for computing the optimal value function $V^*$, and/or the optimal policy $\pi^*$ for MDPs. The arsenal of standard methods include value and policy iteration, and linear programming [3]. However, many of these methods are usually ineffective in hybrid factored MDPs: First, even, if the state and the action spaces are defined solely in terms of discrete variables, the running time of these standard solution methods are polynomial in the number of states and actions, *i.e.*, exponential in the dimension of these spaces. Furthermore, in our hybrid setting, the state and action spaces includes continuous components, often leading to optimal value functions or policies with no compact functional representation.

A standard way to solve such complex MDPs is to assume a surrogate value function form with a small set of tunable parameters. The goal of the optimization is then to identify the value of these parameters that best fit the MDP solution. This approximation of the true value function model by a parametric model is best viewed as transformation of a highly complex high dimensional optimization problem to a lower dimensional one. Particularly popular in recent years are approximations based on linear representations of value functions, where the value function $V(\mathbf{x})$ is expressed as a linear combination of $k$ basis functions $f_i(\mathbf{x})$ [1, 13]:

$$V(\mathbf{x}) = \sum_{i=1}^{k} w_i f_i(\mathbf{x}).$$

In the general case, basis functions are defined over complete state space $\mathbf{X}$, but very often they are restricted only to subsets of state variables [1, 12]. The goal of the optimization is to find the best set of weights $\mathbf{w} = (w_1, \ldots, w_k)$.

### 3.1 Formulation

A variety of methods have been proposed for optimizing the weights $\mathbf{w}$ of the linear value function model [3]. In this paper, we build on the increasingly popular approximate linear programming (ALP) formulation that was first proposed by Schweitzer and Seidmann [15] for discrete MDPs. Their method restricts the general LP-based solution formulation for MDPs to consider only value functions in the restricted linear subspace defined by the basis functions.

Our formulation generalizes ALPs to hybrid settings, optimizing $\mathbf{w}$ by solving a convex optimization problem we call *hybrid approximate linear program* (HALP). The HALP is given by:

$$\begin{aligned} \text{minimize}_{\mathbf{w}} \quad & \sum_i w_i \alpha_i \\ \text{subject to:} \quad & \sum_i w_i F_i(\mathbf{x}, \mathbf{a}) - R(\mathbf{x}, \mathbf{a}) \geq 0 \ \ \forall \, \mathbf{x}, \mathbf{a}; \end{aligned} \tag{3}$$

where $\alpha_i$ in the objective function denotes the *basis function relevance weight* that is given by:

$$\alpha_i = \sum_{\mathbf{x}_D} \int_{\mathbf{x}_C} \psi(\mathbf{x}) f_i(\mathbf{x}) d\mathbf{x}_C, \tag{4}$$

where $\psi(\mathbf{x}) > 0$ is a *state relevance density function* such that $\sum_{\mathbf{x}_D} \int_{\mathbf{x}_C} \psi(\mathbf{x}) d\mathbf{x}_C = 1$, allowing us to weigh the quality of our approximation differently for different parts of the state space; and the $F_i(\mathbf{x}, \mathbf{a})$ function in the constraints is given by:

$$F_i(\mathbf{x}, \mathbf{a}) = f_i(\mathbf{x}) - \gamma \sum_{\mathbf{x}'_D} \int_{\mathbf{x}'_C} p(\mathbf{x}' \mid \mathbf{x}, \mathbf{a}) f_i(\mathbf{x}') d\mathbf{x}'_C. \tag{5}$$

The weights $\mathbf{w}$ optimized in the HALP reduce the complexity of original value optimization problem; instead of optimal values for all possible states, only $k$ weights need to be found. Note that this formulation reduces to the standard discrete-case ALP [15, 9, 6, 14] if the state space $\mathbf{x}$ is discrete, or to the continuous ALP [11] if the state space is continuous.

A number of concerns arise in context of the HALP approximation. First, the formulation of the HALP appears to be arbitrary, that is, it is not immediately clear how it relates to the original hybrid MDP problem. Second, the HALP approximation for the hybrid MDP involves complex integrals that must be evaluated. Third, the number of constraints defining the LP is exponential if the state and action spaces are discrete and infinite if any of the spaces involves continuous components. In the following text we address and provide solutions for each of these concerns.

### 3.2 Theoretical analysis

Our theoretical analysis of the quality of the solution obtained by the HALP formulation follows similar ideas to those used by de Farias and Van Roy to analyze the discrete case [6]. They note that the approximate formulation cannot guarantee an uniformly good approximation of the optimal value function over the whole state space. To address this issue, they define a *Lyapunov function* that weighs states appropriately: a Lyapunov function $L(\mathbf{x}) = \sum_i w_i^L f_i(\mathbf{x})$ with contraction factor $\kappa \in (0, 1)$ for the transition model $P_\pi$, is a strictly positive function such that:

$$\kappa L(\mathbf{x}) \geq \gamma \sum_{\mathbf{x}'_D} \int_{\mathbf{x}'_C} P_\pi(\mathbf{x}' \mid \mathbf{x}) L(\mathbf{x}') d\mathbf{x}'_C. \tag{6}$$

Using this definition, we can extend their result to the hybrid setting:

**Proposition 1** *Let $\mathbf{w}^*$ be an optimal solution to the HALP in Equation (3), then, for any Lyapunov function $L(\mathbf{x})$, we have that:*

$$\|V^* - H\mathbf{w}^*\|_{1,\psi} \leq \frac{2\psi^\intercal L}{1-\kappa} \min_{\mathbf{w}} \|V^* - H\mathbf{w}\|_{\infty, 1/L},$$

*where $H\mathbf{w}$ represents the function $\sum_i w_i f_i(\cdot)$, the $\mathcal{L}_1$ norm weighted by $\psi$ i given by $\|\cdot\|_{1,\psi}$, and $\|\cdot\|_{\infty,1/L}$ is the max-norm weighted by $1/L$.*

**Proof:** *The proof of this result for the hybrid setting follows the outline of the proof of de Farias and Van Roy's Theorem 4.2 [6] for the discrete case.* ■

The main intuition behind this result is that, if our basis functions can approximate the optimal value function in the states emphasized by the Lyapunov function, then $\min_{\mathbf{w}} \|V^* - H\mathbf{w}\|_{\infty,1/L}$ will be low, and the HALP will provide a good approximate solution. We refer the reader to [6] for further discussions.

# 4 Factored HALP

Factored MDP models offer, in addition to structured parameterizations of the process, an opportunity to solve the problem more efficiently. The opportunity stems from the structure of constraint definitions that decompose over state and action subspaces. This is the direct consequence of: (1) factorizations, (2) the presence of local of transition probabilities, and (3) basis functions defined over small state subspaces. This section describes how these properties allow us to compute the factors in the HALP efficiently.

## 4.1 Factored hybrid basis function representation

Koller and Parr [12] show that basis functions with limited scope, that is, each $f_i(\mathbf{x}_i)$ is restricted to depend only on the variables $\mathbf{X}_i \subseteq \mathbf{X}$, provide the basis for efficient approximations in the context of discrete factored MDPs. An important issue in hybrid settings is that the problem formulation incorporates integrals, which may not be computable. Hauskrecht and Kveton [11] propose conjugate transition model and basis function classes that lead to closed-form solutions of all integrals in the strict continuous case. The matching pairs include transitions based on beta or mixture of betas densities, where beta density is defined as in Section 2, and polynomial basis functions. Specifically, each basis function is defined as a product of factors:

$$f_i(\mathbf{x}_i) = \prod_{x_j \in \mathbf{x}_i} x_j^{m_{j,i}}, \tag{7}$$

where $m_{j,i}$ is the degree of the polynomial for variable $X_j$ in basis function $i$.

In our hybrid setting, we must allow basis functions to have both discrete and continuous components. We define each function $f_i(\mathbf{x}_i)$, where $\mathbf{x}_i$ has discrete components $\mathbf{x}_{i_D}$ and continuous components $\mathbf{x}_{i_C}$ by the product of two factors:

$$f_i(\mathbf{x}_i) = f_{i_D}(\mathbf{x}_{i_D}) f_{i_C}(\mathbf{x}_{i_C}), \tag{8}$$

where $f_{i_C}(\mathbf{x}_{i_C})$ takes the form of polynomials over the variables in $\mathbf{X}_{i_C}$, as in Equation (7), and $f_{i_D}(\mathbf{x}_{i_D})$ is an arbitrary function over the discrete variables $\mathbf{X}_{i_D}$. This basis function representation gives us very high flexibility over the class of possible basis functions, but also leads to an efficient solution algorithm for the hybrid planning problem.

## 4.2 Hybrid backprojections

Our first task is to compute the differences in Equation (5) that appear in the constraints for the HALP formulation. Each $F_i(\mathbf{x}, \mathbf{a})$ is the difference between basis function $f_i(\mathbf{x})$ and the discounted *backprojection*, $\gamma g_i(\mathbf{x}, \mathbf{a})$, of this basis function, given by:

$$g_i(\mathbf{x}, \mathbf{a}) = \sum_{\mathbf{x}'_D} \int_{\mathbf{x}'_C} p(\mathbf{x}' \mid \mathbf{x}, \mathbf{a}) f_i(\mathbf{x}') d\mathbf{x}'_C.$$

This backprojection requires us to compute a sum over the exponential number of discrete states $\mathbf{X}'_D$, and integrals over the continuous states $\mathbf{X}'_C$.

For discrete settings, Koller and Parr [12] show that, for basis functions with restricted scope these backprojections can be computed efficiently. Hauskrecht and Kveton [11] extend this idea by showing that for polynomial basis functions and beta transition models, the integrals can be computed in closed-form.

In hybrid settings, we combine these two ideas in a straightforward manner. Using our basis function representation from Equation (8), we can redefine the backprojection:

$$\begin{aligned} g_i(\mathbf{x}, \mathbf{a}) &= g_{i_D}(\mathbf{x}, \mathbf{a}) g_{i_C}(\mathbf{x}, \mathbf{a}), \qquad (9) \\ &= \Big(\sum_{\mathbf{x}'_{i_D}} p(\mathbf{x}'_{i_D} \mid \mathbf{x}, \mathbf{a}) f_{i_D}(\mathbf{x}'_{i_D})\Big) \\ &\qquad \Big(\int_{\mathbf{x}'_{i_C}} p(\mathbf{x}'_{i_C} \mid \mathbf{x}, \mathbf{a}) f_{i_C}(\mathbf{x}'_{i_C}) d\mathbf{x}'_{i_C}\Big). \end{aligned}$$

Note that $g_{i_D}(\mathbf{x}, \mathbf{a})$ is the backprojection of a discrete basis function and $g_{i_C}(\mathbf{x}, \mathbf{a})$ is the backprojection of a continuous basis function. Thus, $g_i(\mathbf{x}, \mathbf{a})$ can be computed by backprojecting each term separately and then multiplying the results. We review these constructions here for completeness.

Let us first consider the backprojection $g_{i_D}(\mathbf{x}, \mathbf{a})$ of the discrete component of the basis function. Note that the scope of this function is restricted to $\mathsf{Par}(\mathbf{X}'_{i_D})$, the parents of the variables $\mathbf{X}'_{i_D}$ in our DBN. Thus, if $f_{i_D}$ is restricted to a small scope, then $g_{i_D}$ will also have restricted scope. If $\mathsf{Par}(\mathbf{X}'_{i_D})$ only includes discrete variables, then we can define $g_{i_D}(\mathbf{x}, \mathbf{a})$ as a table with an entry for each assignment $\mathrm{Dom}(\mathsf{Par}(\mathbf{X}'_{i_D}))$. If some of these parent variables are continuous, we can define $g_{i_D}$ by the combination of a tabular representation with the product of discriminant functions, such as those shown in Equation (1).

For the backprojection $g_{i_C}(\mathbf{x}, \mathbf{a})$ of the continuous component of the basis function, we use the closed-form formula for the integral of a beta distribution with a polynomial basis function described by Hauskrecht and Kveton [11]:

$$g_{i_C}(\mathbf{x}, \mathbf{a}) = \prod_{j:\ X'_j \in \mathbf{X}'_{i_C}} \frac{\Gamma(h_j^1(\mathbf{x}, \mathbf{a}) + h_j^2(\mathbf{x}, \mathbf{a}))\Gamma(h_j^1(\mathbf{x}, \mathbf{a}) + m_{j,i})}{\Gamma(h_j^1(\mathbf{x}, \mathbf{a}) + h_j^2(\mathbf{x}, \mathbf{a}) + m_{j,i})\Gamma(h_j^1(\mathbf{x}, \mathbf{a}))},$$

where $\Gamma(.)$ is a gamma function. Note that the scope of $g_{i_C}(\mathbf{x}, \mathbf{a})$ is restricted to $\mathsf{Par}(\mathbf{X}'_{i_C})$. Therefore, we can compute both $g_{i_D}(\mathbf{x}, \mathbf{a})$ and $g_{i_C}(\mathbf{x}, \mathbf{a})$ in closed-form, yielding a closed-form representation for each $F_i(\mathbf{x}, \mathbf{a})$ in the HALP formulation; finally, obtaining a closed-form formula for each constraint as the sum of restricted scope functions.

## 4.3 Hybrid relevance weights

As with backprojections, computing the basis function relevance weights $\alpha_i$s in Equation (4) that appear in the objective function of the HALP formulation require us to solve exponentially-large sums and complex integrals. Guestrin *et al.* [9, 10] show that if the state relevance density $\psi(\mathbf{x})$ is represented in a factorized fashion, these weights can be computed efficiently. In this section, we show that these ideas also extend for continuous and hybrid settings.

In hybrid settings, we can decompose the computation of $\alpha_i$ in a similar fashion to the backprojections in Equation (9):

$$\begin{aligned}\alpha_i &= \alpha_{i_D}\alpha_{i_C}, \\ &= \left(\sum_{\mathbf{x}_{i_D}} \psi(\mathbf{x}_{i_D}) f_{i_D}(\mathbf{x}_{i_D})\right) \\ &\quad \left(\int_{\mathbf{x}_{i_C}} \psi(\mathbf{x}_{i_C}) f_{i_C}(\mathbf{x}_{i_C}) d\mathbf{x}_{i_C}\right),\end{aligned} \tag{10}$$

where $\psi(\mathbf{x}_{i_D})$ is the marginal of the density $\psi(\mathbf{x})$ to the discrete variables $\mathbf{X}_{i_D}$, and $\psi(\mathbf{x}_{i_C})$ is the marginal to the continuous variables $\mathbf{X}_{i_D}$. As discussed by Guestrin *et al.* [9, 10], we can compute these marginal densities efficiently, for example, by representing $\psi(\mathbf{x})$ as the product of marginals, or as a Bayesian network. Using these marginals, we can compute $\alpha_{i_D}$ and $\alpha_{i_C}$ efficiently, as we show in this section.

Given the marginal $\psi(\mathbf{x}_{i_D})$, the discrete weight $\alpha_{i_D}$ can be obtained by simply enumerating the discrete states in $\mathrm{Dom}(\mathbf{X}_{i_D})$, and using the sum in Equation (10). If each $f_{i_D}(\mathbf{x}_{i_D})$ has scope restricted to a small set of variables, then the number of joint states $|\mathrm{Dom}(\mathbf{X}_{i_D})|$ is relatively small, and $\alpha_{i_D}$ can be computed efficiently.

The second term $\alpha_{i_C}$ from the continuous part of the basis function representation requires us to compute integrals that may be difficult to obtain in closed-form. Note that $\alpha_{i_C}$ is the expectation of $f_{i_C}(\mathbf{x}_{i_C})$ with respect to the density $\psi(\mathbf{x}_{i_C})$:

$$\alpha_{i_C} = \int_{\mathbf{x}_{i_C}} \psi(\mathbf{x}_{i_C}) f_{i_C}(\mathbf{x}_{i_C}) d\mathbf{x}_{i_C} = E_\psi\left[f_{i_C}(\mathbf{x}_{i_C})\right].$$

Note further that the continuous part of our basis functions take the form of polynomials, as in Equation (7). Thus, $E_\psi\left[f_{i_C}(\mathbf{x}_{i_C})\right]$ is simply one of the moments of the density $\psi(\mathbf{x}_{i_C})$. For many densities, including uniform, mixture of Betas, etc, these moments can be obtained in closed-form. For beta densities, for example, the computation is similar to that used for backprojections, presented in the previous section.

## 5 Factored $\varepsilon$-HALP formulation

Despite decompositions, and the closed-form representation for the objective function coefficients and constraints, factored HALPs remain hard to solve. The issue is that, in the general case, the HALP formulation includes one constraint for each joint state $\mathbf{x}$ and action $\mathbf{a}$. The discrete components of these spaces lead to exponentially-many constraints, while the continuous component leads to an uncountably infinite constraint set. To address this problem we propose to transform the factored HALP into $\varepsilon$-HALP, an approximation of the factored HALP with a finite number of constraints.

The $\varepsilon$-HALP relies on the $\varepsilon$ coverage of the constraint space. In the $\varepsilon$-coverage each continuous (state or action) variable is discretized into $\frac{1}{2\varepsilon}+1$ equally spaced values. The discretization induces a multidimensional grid $G$, such that any point in $[0,1]^d$ is at most $\varepsilon$ far from a point in $G$ under the max-norm.

If we directly enumerate each state and action configuration of the $\varepsilon$-HALP we obtain an LP with exponentially-many constraints. However, not all these constraints define the solution and need to be enumerated. This is the same setting as the factored LP decomposition of Guestrin *et al.* [8]. We can use the same technique to decompose our $\varepsilon$-HALP into an equivalent LP with exponentially-fewer constraints. The complexity of this new problem will only be exponentially in the tree-width of a cost network formed by the restricted scope functions in our LP, rather than in the complete set of variables [8, 10]. Alternatively we can also apply the approach by Schuurmans and Patrascu [14] that incrementally builds the set of constraints using a constraint generation heuristic and often performs well in practice. Figure 1 summarizes the main steps of the algorithm to solve the hybrid factored MDP via $\varepsilon$-HALP approximation.

### $\varepsilon$-HALP algorithm

**Inputs:**
$\varepsilon$, HMDP, $f_1, f_2, \cdots f_k$.
**Outputs:**
Basis function weights $\mathbf{w} = w_1, w_2, \cdots w_k$

1. Discretize every continuous state variable $X_u \in [0,1]$ and action variable $A_u \in [0,1]$ using $(\frac{1}{2\varepsilon}+1)$ equally spaced values.
2. For each $F_i(\mathbf{x},\mathbf{a})$ and $R_j(\mathbf{x},\mathbf{a})$ identify subsets $\mathbf{X}_i$ and $\mathbf{A}_i$ ($\mathbf{X}_j$ and $\mathbf{A}_j$) the functions $F_i$ and $R_j$ depend on.
3. Compute $F_i(\mathbf{x}_i,\mathbf{a}_i)$ and $R_j(\mathbf{x}_j,\mathbf{a}_j)$for all possible configurations of values of variables in $\mathbf{X}_i$ and $\mathbf{A}_i$ ($\mathbf{X}_j$ and $\mathbf{A}_j$), for continuous variables use their discretizations.
4. Calculate basis state relevance weights $\alpha_i$.
5. Use the ALP algorithm by Guestrin et al [8] or Schuurmans and Patrascu [14] for factored discrete-valued variables to solve for the optimal weights $\mathbf{w}$.

Figure 1: A summary of the $\varepsilon$-HALP algorithm.

The $\varepsilon$-HALP offers an efficient approximation of a hybrid factored MDP; however, it is unclear how the discretization affects the quality of the approximation. Most discretization approaches require an exponential number of points for a fixed approximation level. In the remainder of this section, we provide a proof that exploits factorization structure to show that our $\varepsilon$-HALP provides a polynomial approximation of the continuous HALP formulation.

### 5.1 Bound on the quality of $\varepsilon$-HALP

A solution to the $\varepsilon$-HALP will usually violate some of the constraints in the original HALP formulation. We show that if these constraints are violated by a small amount, then the $\varepsilon$-HALP solution is near optimal.

Let us first define the degree to which a relaxed HALP, that is, a HALP defined over a finite subset constraints, violates the complete set of constraints.

**Definition 1** *A set of weights* $\mathbf{w}$ *is* $\delta$-infeasible *if:*

$$\sum_i w_i F_i(\mathbf{x},\mathbf{a}) - R(\mathbf{x},\mathbf{a}) \geq -\delta, \ \forall \mathbf{x},\mathbf{a}. \ \blacksquare$$

Now we are ready to show that, if the solution to the relaxed HALP is $\delta$-infeasible, then the quality of the approximation obtained from the relaxed HALP is close to the one in the complete HALP.

**Proposition 2** *Let $\mathbf{w}^*$ be any optimal solution to the complete HALP in Equation (3), and $\widehat{\mathbf{w}}$ be any optimal solution to a relaxed HALP, such that $\widehat{\mathbf{w}}$ is $\delta$-infeasible, then:*

$$\|V^* - H\widehat{\mathbf{w}}\|_{1,\psi} \leq \|V^* - H\mathbf{w}^*\|_{1,\psi} + 2\frac{\delta}{1-\gamma}.$$

**Proof:** First, by monotonicity of the Bellman operator, any feasible solution $\mathbf{w}$ in the complete HALP satisfies:

$$\sum_i w_i f_i(\mathbf{x}) \geq V^*(\mathbf{x}). \tag{11}$$

Using this fact, we have that:

$$\begin{aligned} \|H\mathbf{w}^* - V^*\|_{1,\psi} &= \psi^\intercal \left|H\mathbf{w}^* - V^*\right|, \\ &= \psi^\intercal \left(H\mathbf{w}^* - V^*\right), \\ &= \psi^\intercal H\mathbf{w}^* - \psi^\intercal V^*. \end{aligned} \tag{12}$$

Next, note that the constraints in the relaxed HALP are a subset of those in the complete HALP. Thus, $\mathbf{w}^*$ is feasible for the relaxed HALP, and we have that:

$$\psi^\intercal H\mathbf{w}^* \geq \psi^\intercal H\widehat{\mathbf{w}}. \tag{13}$$

Now, note that if $\widehat{\mathbf{w}}$ is $\delta$-infeasible in the complete HALP, then if we add $\frac{\delta}{1-\gamma}$ to $H\widehat{\mathbf{w}}$ we obtain a feasible solution to the complete HALP, yielding:

$$\begin{aligned} \left\|H\widehat{\mathbf{w}} + \frac{\delta}{1-\gamma} - V^*\right\|_{1,\psi} &= \psi^\intercal H\widehat{\mathbf{w}} + \frac{\delta}{1-\gamma} - \psi^\intercal V^*, \\ &\leq \psi^\intercal H\mathbf{w}^* + \frac{\delta}{1-\gamma} - \psi^\intercal V^*, \\ &= \|H\mathbf{w}^* - V^*\|_{1,\psi} + \frac{\delta}{1-\gamma}. \end{aligned} \tag{14}$$

The proof is concluded by substituting Equation (14) into the triangle inequality bound:

$$\|H\widehat{\mathbf{w}} - V^*\|_{1,\psi} \leq \left\|H\widehat{\mathbf{w}} + \frac{\delta}{1-\gamma} - V^*\right\|_{1,\psi} + \frac{\delta}{1-\gamma}. \blacksquare$$

The above result can be combined with the result in Section 3 to obtain the bound on the quality of the $\varepsilon$-HALP.

**Theorem 1** *Let $\widehat{\mathbf{w}}$ be any optimal solution to the relaxed $\varepsilon$-HALP satisfying the $\delta$ infeasibility condition. Then, for any Lyapunov function $L(\mathbf{x})$, we have:*

$$\|V^* - H\widehat{\mathbf{w}}\|_{1,\psi} \leq 2\frac{\delta}{1-\gamma} + \frac{2\psi^\intercal L}{1-\kappa}\min_{\mathbf{w}} \|V^* - H\mathbf{w}\|_{\infty,1/L}.$$

**Proof:** Direct combination of Propositions 1, 2. $\blacksquare$

### 5.2 Resolution of the $\varepsilon$ grid

Our bound for relaxed versions on the HALP formulation, presented in the previous section, relies on adding enough constraints to guarantee at most $\delta$-infeasibility. The $\varepsilon$-HALP approximates the constraints in HALP by restricting values of its continuous variables to the $\varepsilon$-grid. In this section, we analyze the relationship between the choice of $\varepsilon$ and the violation level $\delta$, allowing us to choose the appropriate discretization level for a desired approximation error in Theorem 1.

Our condition in Definition 1 can be satisfied by a set constraints $\mathcal{C}$ that ensures a $\delta$ max-norm discretization of $\sum_i \widehat{w}_i F_i(\mathbf{x},\mathbf{a}) - R(\mathbf{x},\mathbf{a})$. In the $\varepsilon$-HALP this condition is met with the $\varepsilon$-grid discretization that assures that for any state-action pair $\mathbf{x},\mathbf{a}$ there exists a pair $\mathbf{x}_G,\mathbf{a}_G$ in the $\varepsilon$-grid such that:

$$\left\|\textstyle\sum_i \widehat{w}_i F_i(\mathbf{x},\mathbf{a}) - R(\mathbf{x},\mathbf{a}) - \sum_i \widehat{w}_i F_i(\mathbf{x}_G,\mathbf{a}_G) - R(\mathbf{x}_G,\mathbf{a}_G)\right\|_\infty \leq \delta.$$

Usually, such bounds are achieved by considering the Lipschitz modulus of the discretized function: Let $h(\mathbf{u})$ be an arbitrary function defined over the continuous subspace $\mathbf{U} \in [0,1]^d$ with a Lipschitz modulus $K$ and let $G$ be an $\varepsilon$-grid discretization of $\mathbf{U}$. Then the $\delta$ max-norm discretization of $h(\mathbf{u})$ can be achieved with a $\varepsilon$-grid with the resolution $\varepsilon \leq \frac{\delta}{K}$. Usually, the Lipschitz modulus of a function rapidly increases with dimension $d$, thus requiring additional points for a desired discretization level.

Each constraint in the $\varepsilon$-HALP is defined in terms of a sum of functions: $\sum_i \widehat{w}_i F_i(\mathbf{x},\mathbf{a}) - \sum_j R(\mathbf{x},\mathbf{a})$, where each function depends only on a small number of variables (and thus has a small dimension). Therefore, instead of using a global Lipschitz constant $K$ for the complete expression we can express the relation in between the factor $\delta$ and $\varepsilon$ in terms of the Lipschitz constants of individual functions, exploiting the factorization structure. In particular, let $K_{max}$ be the worst-case Lipschitz constant over both the reward functions $R_j(\mathbf{x},\mathbf{a})$ and $w_i F_i(\mathbf{x},\mathbf{a})$. To guarantee that $K_{max}$ is bounded, we must bound the magnitude of $\widehat{w}_i$. Typically, if the basis functions have unit magnitude, the $\widehat{w}_i$ will be bounded $R_{max}/(1-\gamma)$. Here, we can define $K_{max}$ to be the maximum of the Lipschitz constants of the reward functions and of $R_{max}/(1-\gamma)$ times the constant for each $F_i(\mathbf{x},\mathbf{a})$. By choosing an $\varepsilon$ discretization of only:

$$\varepsilon \leq \frac{\delta}{MK_{max}},$$

where $M$ is the number of functions, we guarantee the condition of Theorem 1 for a violation of $\delta$.

## 6 Experiments

This section presents an empirical evaluation of our approach, demonstrating the quality of the approximation and the scaling properties.

### 6.1 Irrigation network example

An irrigation system consists of a network of irrigation channels that are connected by regulation devices (see Figure 2a).

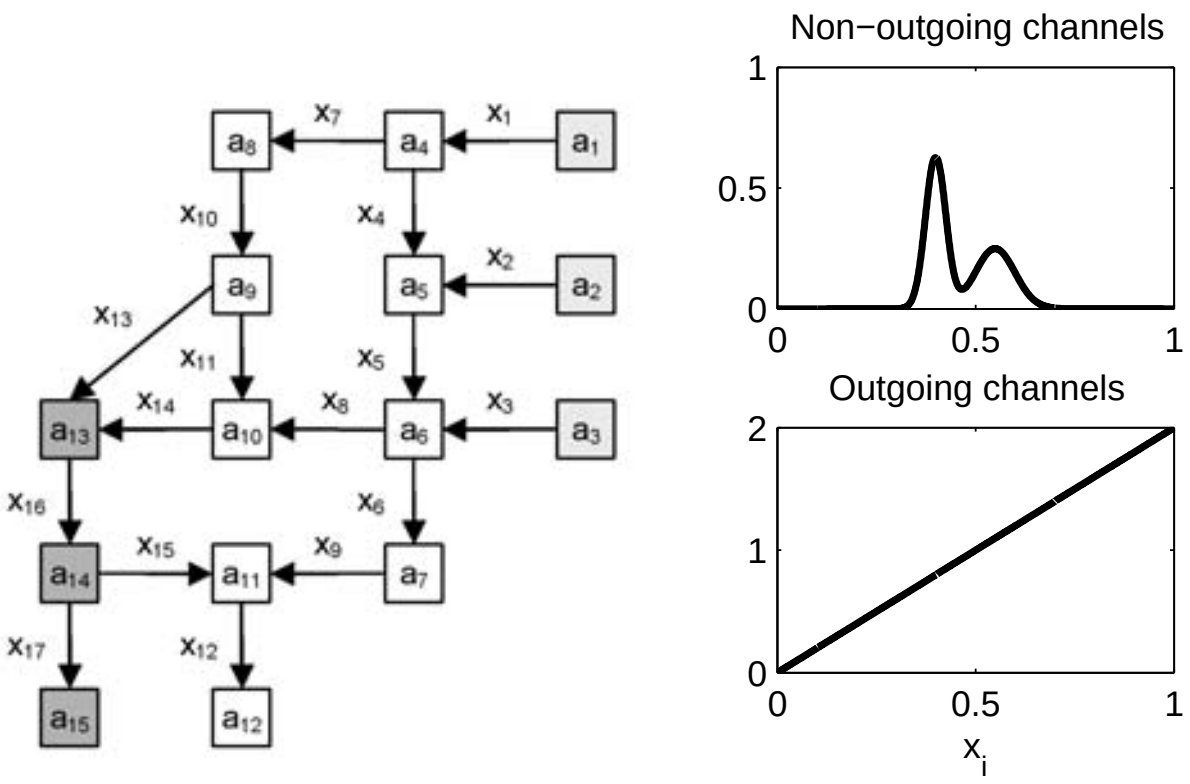

**(a)** **(b)**

Figure 2: **a.** The topology of an irrigation system. Irrigation channels are represented by links $x_i$ and water regulation devices are marked by rectangles $a_i$. Input and output regulation devices are shown in light and dark gray colors. **b.** Reward functions for the amount of water $x_i$ in the $i$th irrigation channel.

Regulation devices are used to regulate the amount of water in the channels, which is achieved by pumping the water from one of the channels to another one. The goal of the operator of the irrigation system is to keep the amount of water in all channels on an optimal level (determined by the type of planted crops, etc.), by manipulation of regulation devices.

Figure 2a illustrates the topology of channels and regulation devices for one of the irrigation systems used in the experiments. To keep problem formulation simple, we adopt several simplifying assumptions: all channels are of the same size, water flows are oriented, and the control structures operate in discrete modes.

The irrigation system can be formalized as a hybrid MDP, and the optimal behavior of the operator can be found as the optimal control policy for the MDP. The amount of water in the $i$th channel is naturally represented by a continuous state factor $x_i \in [0, 1]$. Each regulation device can operate in multiple modes: the water can be pumped in between any pair of incoming and outgoing channel. These options are represented by discrete action variables $a_i$, one variable per regulation device. The input and output regulation devices (devices with no incoming or no outgoing channels) are special and continuously pump the water in or out of the irrigation system. Transition functions are defined as beta densities that represent water flows depending on the operating modes of the regulation devices. Reward function reflects our preference for the amount of water in different channels (Figure 2b). The reward function is factorized along channels, defined by a linear reward function for the outgoing channels, and a mixture of Gaussians for all other channels. The discount factor is $\gamma = 0.95$. To approximate the optimal value function, a combination of linear and piecewise linear feature functions is used at every channel (Figure 3).

### 6.2 Experimental results

The objective of the first set of experiments was to compare the quality of solutions obtained by the $\varepsilon$-HALP for varying grid resolutions $\varepsilon$ against other techniques for policy generation and to illustrate time (in seconds) needed to solve the $\varepsilon$-HALP problem. All experiments are performed on the irrigation network from Figure 2a with 17 dimensional state space and 15 dimensional action space. The results are presented in Figure 4. The quality of policies is measured in terms of the average reward that is obtained via Monte Carlo simulations of the policy on 100 state-action trajectories, each of 100 steps. To assure the fairness of the comparison, the set of initial states is kept fixed across experiments.

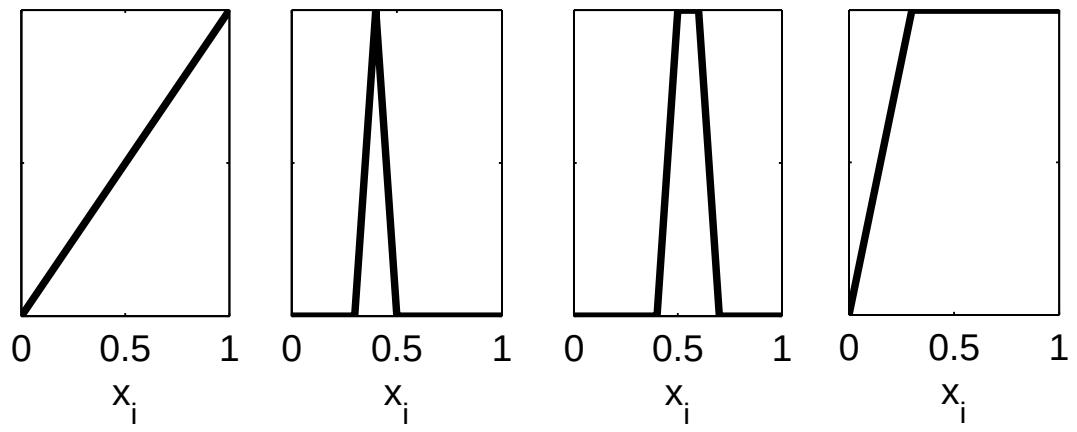

Figure 3: Feature functions for the amount of water $x_i$ in the $i$th irrigation channel.

| **$\varepsilon$-HALP** | | | | **Alternative solution** | | |
|---|---|---|---|---|---|---|
| $\varepsilon$ | $\mu$ | $\sigma$ | Time[s] | Method | $\mu$ | $\sigma$ |
| 1 | 42.8 | 3.0 | 2 | Random | 35.9 | 2.7 |
| 1/2 | 60.3 | 3.0 | 21 | Local | 55.4 | 2.5 |
| 1/4 | 61.9 | 2.9 | 184 | Global 1 | 60.4 | 3.0 |
| 1/8 | 72.2 | 3.5 | 1068 | Global 4 | 66.0 | 3.6 |
| 1/16 | 73.8 | 3.0 | 13219 | Global 16 | 68.2 | 3.2 |

Figure 4: Results of the experiments for the irrigation system in Figure 2a. The quality of found policies is measured by the average reward $\mu$ for 100 state-action trajectories. $\sigma$ denotes the standard deviation of reward values.

Three alternative solutions are used in the comparison: random policy, local heuristic, and global heuristic. The random policy operates regulation devices randomly and serves as a baseline solution. The local heuristic optimizes the one-step expected reward for every regulation device locally, while ignoring all other devices. Finally, the global heuristic attempts to optimize one-step expected reward for all regulatory devices together. The parameter of the global heuristic is the number of trials used to estimate the global one-step reward. All heuristic solutions were applied in the on-line mode; thus, their solution times are not included in Figure 4. The results show that the $\varepsilon$-HALP is able to solve a very complex optimization problem relatively quickly and outperform strawman heuristic methods in terms of the quality of their solutions.

### 6.3 Scale-up study

The second set of experiments focuses on the scale-up potential of $\varepsilon$-HALP method with respect to the complexity of the model. The experiments are performed for $n$-ring and $n$-ring-of-rings topologies (see Figure 5a). The results, summarized in Figure 5b, show several important trends: (1) the quality of the policy for the $\varepsilon$-HALP improves with higher grid resolution $\varepsilon$, (2) the running time of the method grows exponentially with the grid resolution, and (3) the increase in the running time of the method for topologies of increased complexity is mild and far from exponential in the number of variables $n$. Graphical examples of each of these trends are given in Figures 5c, 5d, and 5e. In addition to the running time curve, Figure 5e shows a quadratic polynomial fitted to the values for different $n$. This supports our theoretical findings that the running time complexity of the $\varepsilon$-HALP method for an appropriate choice of basis functions does not grow exponentially in the number of variables.

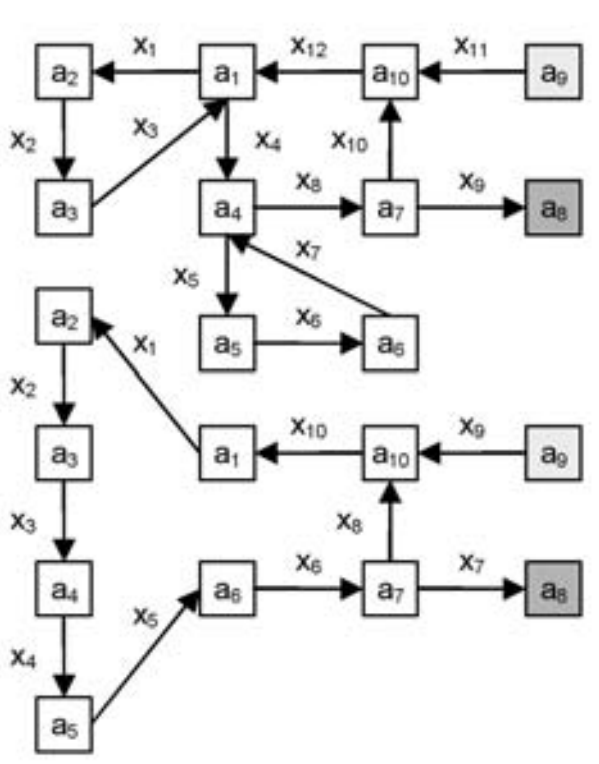

(a)

| $\varepsilon$ | $n$-ring | | | | | | | | | |
|---|---|---|---|---|---|---|---|---|---|---|
| | $n = 6$ | | $n = 9$ | | $n = 12$ | | $n = 15$ | | $n = 18$ | |
| | $\mu$ | Time[s] | $\mu$ | Time[s] | $\mu$ | Time[s] | $\mu$ | Time[s] | $\mu$ | Time[s] |
| 1 | 28.4 | 1 | 37.5 | 1 | 46.9 | 1 | 55.6 | 2 | 64.5 | 3 |
| 1 / 2 | 33.5 | 3 | 43.0 | 5 | 52.6 | 9 | 62.9 | 17 | 72.1 | 28 |
| 1 / 4 | 35.1 | 11 | 45.2 | 21 | 54.2 | 43 | 64.2 | 63 | 74.5 | 85 |
| 1 / 8 | 40.1 | 46 | 51.4 | 85 | 62.2 | 118 | 73.2 | 168 | 84.9 | 193 |
| 1 / 16 | 40.4 | 331 | 51.8 | 519 | 63.7 | 709 | 75.5 | 963 | 86.8 | 1285 |
| $\varepsilon$ | $n$-ring-of-rings | | | | | | | | | |
| | $n = 6$ | | $n = 9$ | | $n = 12$ | | $n = 15$ | | $n = 18$ | |
| | $\mu$ | Time[s] | $\mu$ | Time[s] | $\mu$ | Time[s] | $\mu$ | Time[s] | $\mu$ | Time[s] |
| 1 | 14.8 | 1 | 16.2 | 2 | 17.5 | 4 | 18.5 | 5 | 19.7 | 6 |
| 1 / 2 | 38.6 | 12 | 50.5 | 25 | 44 | 103 | 75.8 | 69 | 87.6 | 107 |
| 1 / 4 | 40.1 | 82 | 53.6 | 184 | 66.7 | 345 | 79.0 | 590 | 93.1 | 861 |
| 1 / 8 | 48.0 | 581 | 62.4 | 1250 | 76.1 | 2367 | 90.5 | 3977 | 104.5 | 6377 |
| 1 / 16 | 47.1 | 4736 | 62.3 | 11369 | 77.6 | 22699 | 92.4 | 35281 | 107.8 | 53600 |

(b)

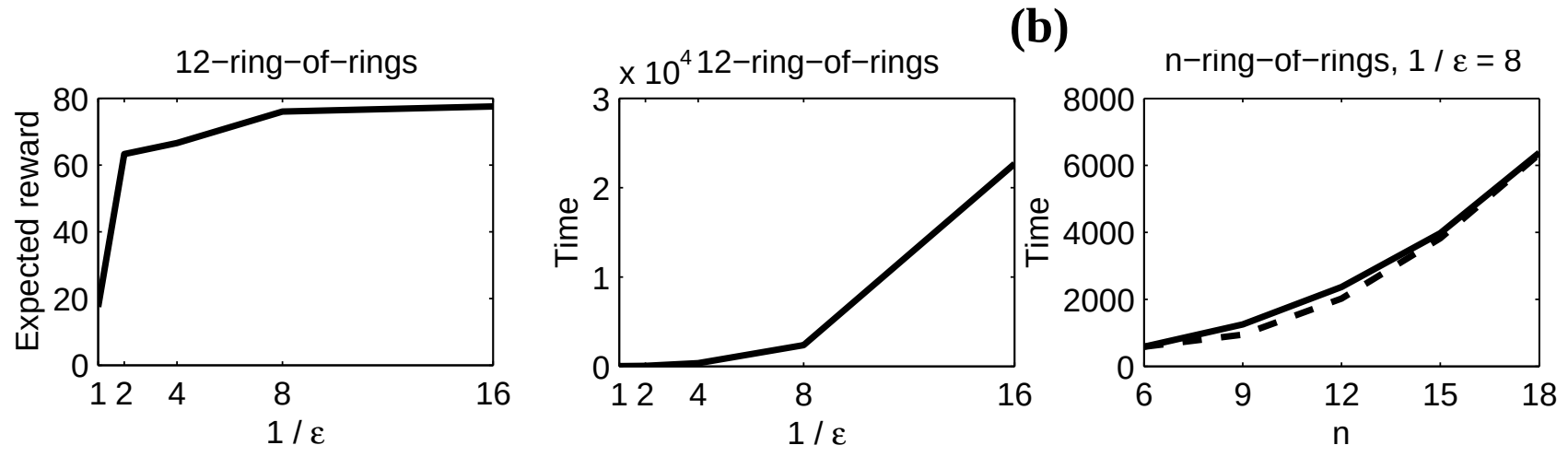

(c) (d) (e)

Figure 5: **a.** Two irrigation network topologies used in the scale-up experiments: $n$-ring-of-rings (shown for $n = 6$) and $n$-ring (shown for $n = 6$). **b.** Average rewards and policy computation times for different $\varepsilon$ and various networks architectures. **c.** Average reward as a function of grid resolution $\varepsilon$. **d.** Time complexity as a function of grid resolution $\varepsilon$. **e.** Time complexity (solid line) as a function of different network sizes $n$. The quadratic approximation of the time complexity is plotted as dashed line.

## 7 Conclusions

We present the first framework that can exploit problem structure for modeling and approximately solving hybrid problems efficiently. We provide bounds on the quality of the solutions obtained by our HALP formulation with respect to the best approximation in our basis function class. This HALP formulation can be closely approximated by the (relaxed) $\varepsilon$-HALP, if the resulting solution is near feasible in the original HALP formulation. Although we would typically require an exponentially-large discretization to guarantee this near feasibility, we provide an algorithm that can efficiently generate an equivalent guarantee with an exponentially-smaller discretization. When combined, these theoretical results lead to a practical algorithm that we have successfully demonstrated on a set of control problems with up to 28-dimensional continuous state space and 22-dimensional action space.

The techniques presented in this paper directly generalize to collaborative multiagent settings, where each agent is responsible for one of the action variables, and they must coordinate to maximize the total reward. The off-line planning stage of our algorithm remains unchanged. However, in the on-line action selection phase, at every time step, the agents must coordinate to choose the action that jointly maximizes the expected value for the current state. We can achieve this by extending the *coordination graph* algorithm of Guestrin *et al.* [9] to our hybrid setting with our factored discretization scheme. The result will be an efficient distribute coordination algorithm that can cope with both continuous and discrete actions.

Many real-world problems involve continuous and discrete elements. We believe that our algorithms and theoretical results will significantly further the applicability of automated planning algorithms to these settings.